\documentclass[letterpaper, 10 pt, conference]{ieeeconf} % TODO use this
\IEEEoverridecommandlockouts                              % This command is only needed if 
\usepackage[whole]{bxcjkjatype} % TODO remove
\usepackage{graphicx}
\usepackage{amsmath}
\usepackage{amssymb}
\usepackage{algorithm}
\usepackage{algpseudocode}
\usepackage{times}
\usepackage{url}
\newcommand{\figref}[1]{Fig.~\ref{figure:#1}}
\newcommand{\tabref}[1]{Table~\ref{table:#1}}

\title{\LARGE \bf
Semantic Scene Difference Detection in Daily Life Patroling by Mobile Robots using Pre-Trained Large-Scale Vision-Language Model
}

\author{Yoshiki Obinata$^{1}$, Kento Kawaharazuka$^{1}$, Naoaki Kanazawa$^{1}$, Naoya Yamaguchi$^{1}$, Naoto Tsukamoto$^{1}$, \\
  Iori Yanokura$^{1}$, Shingo Kitagawa$^{1}$, Koki Shinjo$^{1}$, Kei Okada$^{1}$ and Masayuki Inaba$^{1}$% <-this % stops a space
  \thanks{$^{1}$The authors are with the Department of Mechano-Informatics, Graduate School of Information Science and Technology, The University of Tokyo, 7-3-1 Hongo, Bunkyo-ku, Tokyo, 113-8656, Japan. [obinata, kawaharazuka, kanazawa, yamaguchi, tsukamoto, yanokura, s-kitagawa, shinjo, k-okada, inaba]@jsk.imi.i.u-tokyo.ac.jp
}}

\begin{document}

\maketitle
\thispagestyle{empty}
\pagestyle{empty}

\begin{abstract}
It is important for daily life support robots to detect changes in their environment and perform tasks. In the field of anomaly detection in computer vision, probabilistic and deep learning methods have been used to calculate the image distance. These methods calculate distances by focusing on image pixels. In contrast, this study aims to detect semantic changes in the daily life environment using the current development of large-scale vision-language models. Using its Visual Question Answering (VQA) model, we propose a method to detect semantic changes by applying multiple questions to a reference image and a current image and obtaining answers in the form of sentences. Unlike deep learning-based methods in anomaly detection, this method does not require any training or fine-tuning, is not affected by noise, and is sensitive to semantic state changes in the real world. In our experiments, we demonstrated the effectiveness of this method by applying it to a patrol task in a real-life environment using a mobile robot, Fetch Mobile Manipulator. In the future, it may be possible to add explanatory power to changes in the daily life environment through spoken language.
\end{abstract}

\section{INTRODUCTION}
Robots are becoming capable of performing a variety of life-support behaviors. For robots to spontaneously perform these life-support actions, it is necessary to capture changes in the daily life environment in which the robot is operating.

In the field of computer vision, the task of capturing changes in images includes anomaly detection. Research on anomaly detection has been conducted for many years. Methods based on probabilistic models like Bayesian networks\cite{pearl1985bayesian} and hidden Markov models\cite{baum1966statistical} are used for it, and in recent years, deep learning methods have been used to deal with complex situations. For example, some use Variational Auto Encoder\cite{kingma2013auto}, some use generative models such as GAN\cite{goodfellow2020generative}. These methods require collecting a large amount of data to train the model. In contrast, our study uses language to detect scene difference in daily life environment. Starting with Transformer\cite{vaswani2017attention}, models such as BERT\cite{devlin2018bert}, T5\cite{raffel2020exploring} have demonstrated remarkable performance in language tasks. Language models are also beginning to be introduced into robotics fields\cite{anderson2018vision}\cite{saycan2022arxiv}\cite{kawaharazuka2023vqa}\cite{kawaharazuka2023robotic}\cite{das2018neural}\cite{shah2023lm}\cite{huang2023visual}. In computer vision, large-scale vision-language models trained using these language models have been studied extensively, starting with VQA\cite{antol2015vqa}, and in recent years, trained models capable of performing various vision-language tasks, such as CLIP\cite{radford2021learning}, GLIP\cite{li2021grounded}\cite{zhang2022glipv2}, and OFA\cite{wang2022ofa} have begun to be made available. Unlike image recognition models such as SSD\cite{liu2016ssd} and VGG\cite{simonyan2014very}, these models are difficult to train locally on a small number of GPUs, but they are trained on a large amount of text and image datasets with rich computational resources and have knowledge about the language of human society and images. Therefore, it is possible to accurately extract features from a single image and translate them into language. There are some studies that use pre-trained models for anomaly detection in videos and datasets\cite{10.1007/s10586-021-03439-5}\cite{esmaeilpour2022zero}.

Based on this background, we propose a method to apply a large-scale vision-language model that has already been trained to calculate scene distance in daily life environment for a mobile robot. The large-scale model answers questions about the image captured by the mobile robot's camera and compares the sentences between reference and current images to detect how different the situation is in the same location. This sentence comparison is performed by preparing multiple questions in the VQA task and numerically comparing the answers between reference and current image. This method requires only one reference image for each location and does not require any model re-training.

\section{SCENE DIFFERENCE DETECTION USING PRE-TRAINED VISION LANGUAGE MODELS}
\subsection{Overview of semantic scene difference detection system using pre-trained vision-language model}
\begin{figure}[tb]
  \centering
  \includegraphics[width=\columnwidth]{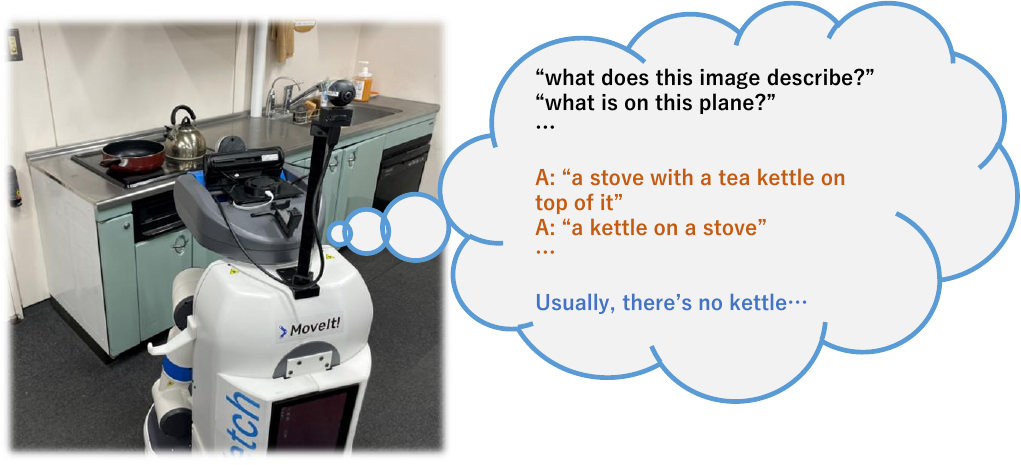}
  % \caption{Anomaly detection using language by a patrol robot. It recognizes what is in front of it in spoken language and identifies the difference in its sentences between normal and current.}
  \caption{Semantic scene difference detection using spoken language by the robot. The robot perceives its daily life environment in the space of spoken language and uses its semantic differential to detect changes.}
  \label{figure:concept}
\end{figure}

\begin{figure*}[t]
  \centering
  \includegraphics[width=0.9\linewidth]{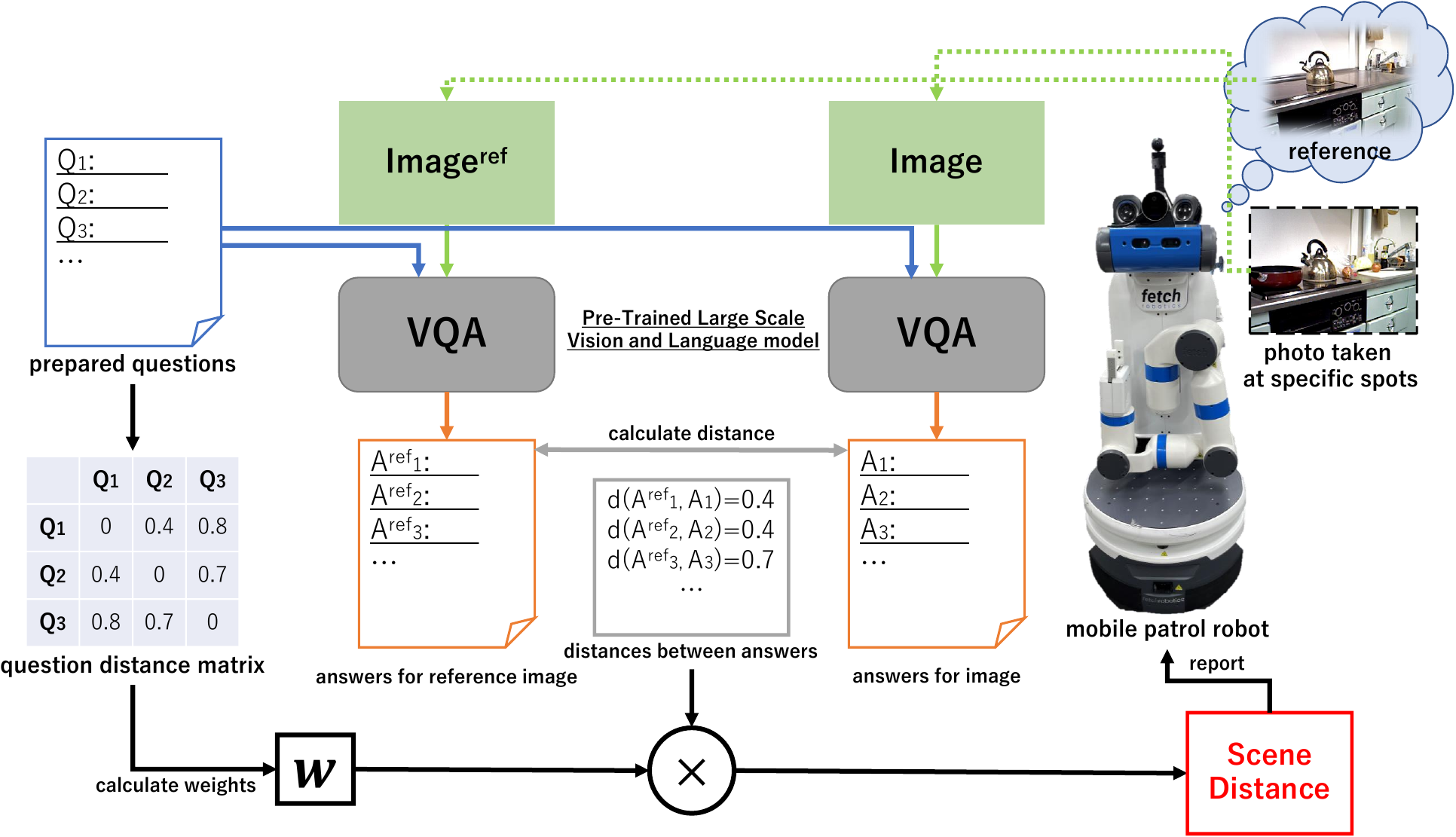}
  \caption{Overall view of semantic scene difference quantifying system using pre-trained vision-language model. The robot uses a large-scale vision-language VQA model to compare reference and captured images, deriving their semantic distance by analyzing the answers to questions about the images. The distance between the answer sentences is calculated using doc2vec, resulting in two vectors whose distance determines the semantic distance between the images. Weights can be given to elements of the vectors to reduce the effect of question proximity.}
  \label{figure:whole-system}
\end{figure*}

The concept of this research is shown in \figref{concept}. Robots can capture semantic changes by comparing situations in their daily life environment through language. It is important to quantify these changes to detect anomalies and initiate tasks in the daily life environment based on these results. \figref{whole-system} shows an overall view of the method used to quantifying its semantic change. The patrol robot compares the reference and captured images and derives their semantic distance. A large-scale vision-language VQA model is given multiple question sentences, a reference image, and a captured image, and the answer sentences for each image are obtained. The distance between the answer sentences corresponding to the same question sentence in the two images is determined by doc2vec, and two vectors of the same dimension as the number of question sentences are derived. By finding the distance between these two vectors, the semantic distance between the two images can be obtained, but by giving appropriate weights to the elements of the vectors, the effect of the semantic proximity of the questions can be reduced.

\subsection{Features of Large-Scale Vision-Language Models in Scene Difference Detection Tasks}

\begin{figure}[t]
  \centering
  \includegraphics[width=\columnwidth]{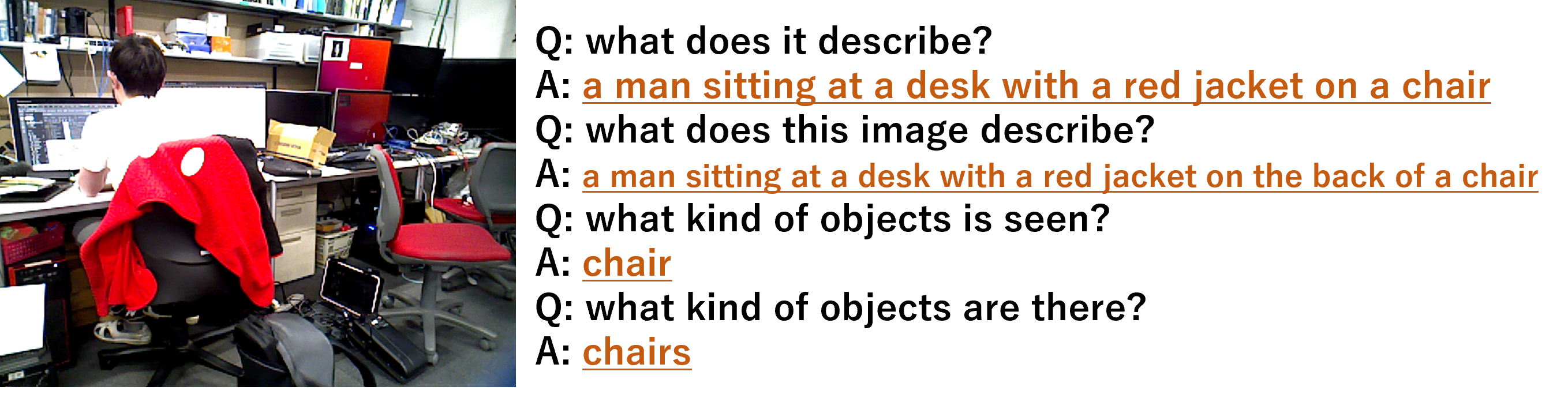}
  \caption{Examples of differences in output sentences depending on the words used. Pronouns and singular and plural forms slightly change the output sentences.}
  \label{figure:caption-expression}
\end{figure}

In this study, we use OFA as a large-scale vision-language model. There are five types of tasks for generating rectangles and text from images: Visual Grounding, Grounded Captioning, Image Text Matching, Image Captioning, and Visual Question Answering. Image Captioning (IC) and Visual Question Answering (VQA) are suitable for explaining the situation, as they allow the user to ask questions and obtain the answers. VQA in OFA sometimes outputs strange sentences like ``prototype prototype prototype of table'', ``messily messy office space'' so we use IC.

The model outputs slightly different answers to questions with similar expressions, as shown in \figref{caption-expression}. Since the output of the model changes depending on the prepositions, pronouns, and verbs used, it makes sense to input a variety of expressions, even for the similarly worded questions.

\subsection{Evaluation of Scene Difference}
\label{subsec:evalanomaly}

We describe a method for calculating the scene difference distance at a patrol location. As shown in \figref{whole-system}, dozens of questions are prepared in advance to ask a large-scale model to describe a situation. Suppose that $m$ questions are prepared, and $m$ answers can be obtained by inputting a single image and $m$ questions to OFA. The questions and answers are transformed into a 768-dimensional vector via all-mpnet-base-v2 \cite{allmpnetbasev2}, a pre-trained language model that further fine-tunes MPNet \cite{song2020mpnet}. Define these as
\begin{equation*}
  \begin{split}
    \boldsymbol{q_{1}}, \boldsymbol{q_{2}}, ..., \boldsymbol{q_{m}} \\
    \boldsymbol{a_{1}}, \boldsymbol{a_{2}}, ..., \boldsymbol{a_{m}}
  \end{split}
\end{equation*}
respectively. Let
\begin{equation*}
  \begin{split}
    \boldsymbol{a^{ref}_{1}}, \boldsymbol{a^{ref}_{2}}, ..., \boldsymbol{a^{ref}_{m}}
  \end{split}
\end{equation*}
be the sentence vectors of the reference responses, and define the Scene Distance $SD$ as
\begin{equation*}
  \begin{split}
    SD = \sum_{k=1}^{m} w_{k}D_{c}(\boldsymbol{a_{k}}, \boldsymbol{a^{ref}_{k}})
  \end{split}
\end{equation*}
using the weights $w_{k}$. $D_{c}$ is the cosine distance, defined by follows
\begin{equation*}
  \begin{split}
    D_{c}(\boldsymbol{p}, \boldsymbol{q}) = 1 - \frac{\boldsymbol{p}\cdot\boldsymbol{q}}{||\boldsymbol{p}||||\boldsymbol{q}||}
  \end{split}
\end{equation*}

Normally, the weights $w_{k}$ is calculated as follows
\begin{equation}
  \label{avg}
  \begin{split}
    w_{k} = \frac{1}{m}
  \end{split}
\end{equation}
However, this method makes it difficult for questions of different types to be reflected in the scene distance. Therefore, we propose the following method to calculate the weights $\boldsymbol{w}$
\begin{equation}
  \label{ours}
  \begin{split}
    M_{rel} & =
    \begin{pmatrix}
      D_{c}(\boldsymbol{q_{1}}, \boldsymbol{q_{1}}) & D_{c}(\boldsymbol{q_{1}}, \boldsymbol{q_{2}}) & \cdots & D_{c}(\boldsymbol{q_{1}}, \boldsymbol{q_{m}})\\
      D_{c}(\boldsymbol{q_{2}}, \boldsymbol{q_{1}}) & D_{c}(\boldsymbol{q_{2}}, \boldsymbol{q_{2}}) & \cdots & D_{c}(\boldsymbol{q_{2}}, \boldsymbol{q_{m}})\\
      \vdots & \vdots & \ddots & \vdots\\
      D_{c}(\boldsymbol{q_{m}}, \boldsymbol{q_{1}}) & D_{c}(\boldsymbol{q_{m}}, \boldsymbol{q_{2}}) & \cdots & D_{c}(\boldsymbol{q_{m}}, \boldsymbol{q_{m}})
    \end{pmatrix}
    \\
    & =
    \begin{pmatrix}
      0 & D_{c}(\boldsymbol{q_{1}}, \boldsymbol{q_{2}}) & \cdots & D_{c}(\boldsymbol{q_{1}}, \boldsymbol{q_{m}})\\
      D_{c}(\boldsymbol{q_{1}}, \boldsymbol{q_{2}}) & 0 & \cdots & D_{c}(\boldsymbol{q_{2}}, \boldsymbol{q_{m}})\\
      \vdots & \vdots & \ddots & \vdots\\
      D_{c}(\boldsymbol{q_{1}}, \boldsymbol{q_{m}}) & D_{c}(\boldsymbol{q_{2}}, \boldsymbol{q_{m}}) & \cdots & 0
    \end{pmatrix}
    \\
    \boldsymbol{v} & = \begin{pmatrix}1 \\ 1 \\ \vdots \\ 1 \end{pmatrix}
    \\
    \boldsymbol{w} & = \frac{M_{rel}\boldsymbol{v}}{||M_{rel}\boldsymbol{v}||}
  \end{split}
\end{equation}
The weights $\boldsymbol{w}$ in (\ref{ours}) have the function of decreasing the contribution of similar meaning questions and increasing the contribution of questions with different meanings from others.

The further apart the meanings of the questions are, the more scene difference can be captured. We define Quality of Questions $QoQ$, a measure of the goodness of the questions being chosen, as
\begin{equation*}
  \begin{split}
    QoQ = \sum_{i, j}{{M_{rel}}_{ij}}
  \end{split}
\end{equation*}
This is the sum of all the elements of $M_{rel}$. The more different the meanings of the questions are from each other, the larger the sum becomes, so the larger $QoQ$ is, the better the questions are selected.

\section{EXPERIMENTS}
\begin{figure}[tb]
  \centering
  \includegraphics[width=0.8\columnwidth]{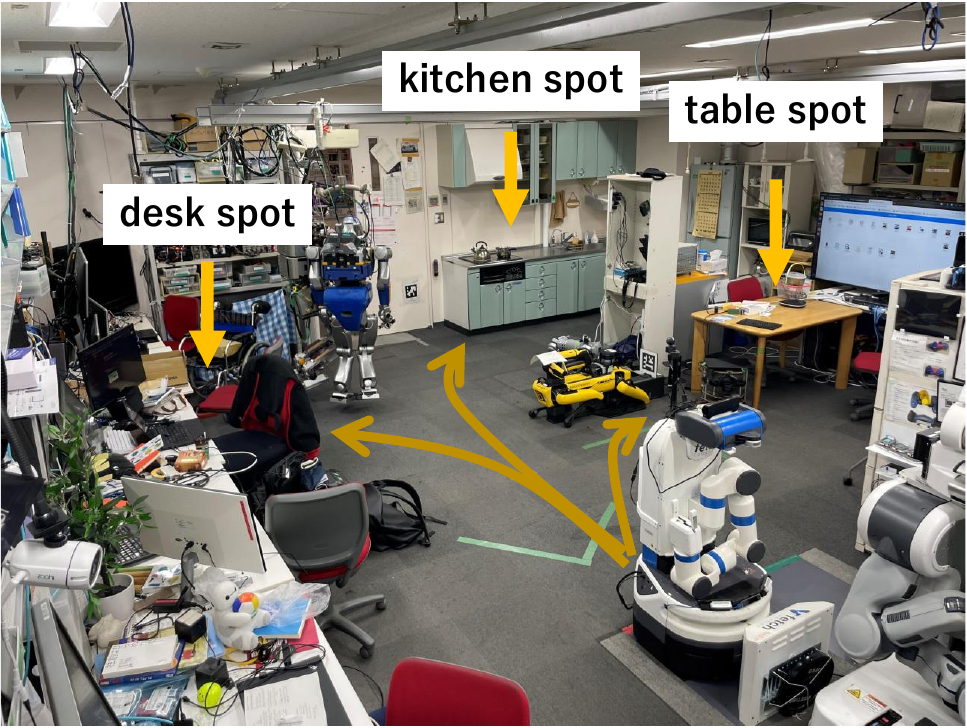}
  \caption{Experimental environment and name of the spot. Fetch Mobile Manipulator navigates to each spot for scene difference detection and points the head camera to the spot to take pictures. Locations where the robot would be expected to perform life support tasks were selected: dining table (table spot), kitchen (kitchen spot), and office desk (desk spot).}
  \label{figure:demo-place}
\end{figure}
\begin{table}[h]
  \caption{Navigation coordinates for each spot}
  \label{table:coords}
  \begin{center}
    \begin{tabular}{|c||c|c|c|}
      \hline
      Spot Name & x[m] & y[m] & yaw[rad] \\
      \hline
      \hline
      table spot & 4.036 & 7.344 & 1.753 \\
      \hline
      kitchen spot & 1.559 & 7.231 & 2.296 \\
      \hline
      desk spot & 4.319 & 6.108 & -2.231 \\
      \hline
    \end{tabular}
  \end{center}
\end{table}
\begin{table}[h]
  \caption{Robot posture for taking a picture at each spot}
  \label{table:pos}
  \begin{center}
    \begin{tabular}{|c||c|c|c|}
      \hline
      Spot Name & torso[mm] & neck yaw[deg] & neck pitch[deg] \\
      \hline
      \hline
      table spot & 21.57 & -2.170 & 19.57 \\
      \hline
      kitchen spot & 21.56 & 3.178 & 16.32 \\
      \hline
      desk spot & 21.58 & -1.116 & 10.83 \\
      \hline
    \end{tabular}
  \end{center}
\end{table}

\begin{figure*}[tb]
  \centering
  \includegraphics[width=0.9\linewidth]{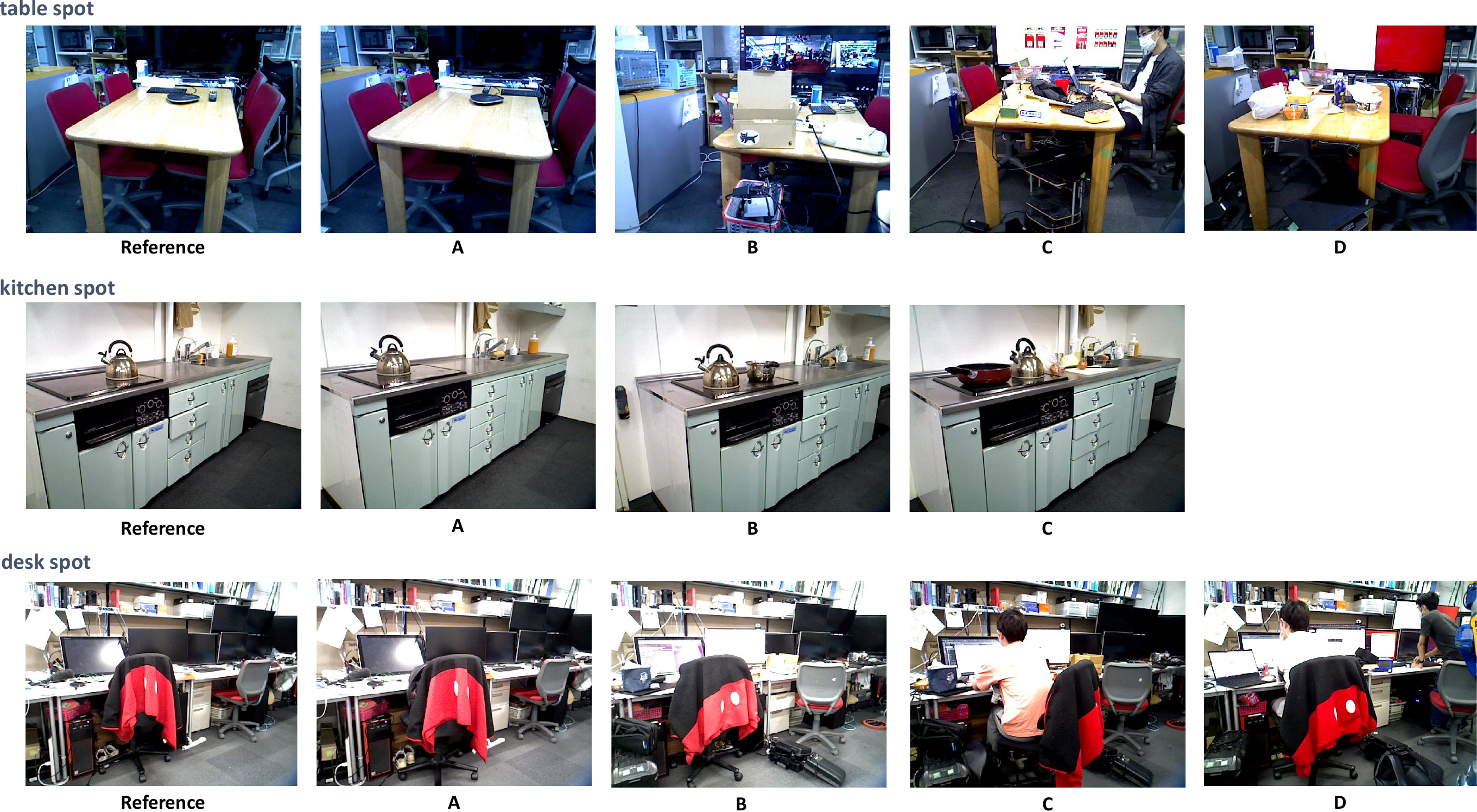}
  \caption{Images of table spot, kitchen spot, and desk spot taken by Fetch Mobile Manipulator. Only one image of the reference state exists for each spot. Images A, B, C, and D taken at different times are compared with this reference image.}
  \label{figure:experiments-photos}
\end{figure*}

\begin{figure}[tb]
  \centering
  \includegraphics[width=0.9\columnwidth]{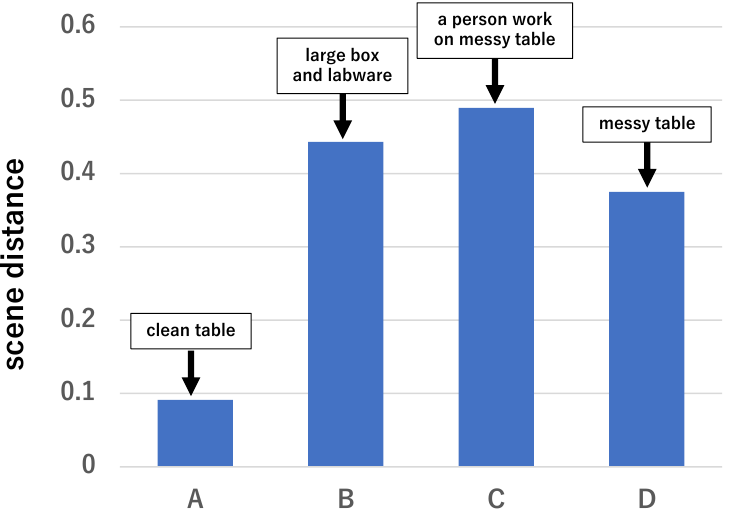}
  \caption{Scene distance and feature of four types of scenes at table spot. The scene distance of image A, which is almost the same situation as the reference, is small, and the scene distances of the other images change with the degree of semantic difference in the situation.}
  \label{figure:table-results}
\end{figure}

\begin{figure}[tb]
  \centering
  \includegraphics[width=0.9\columnwidth]{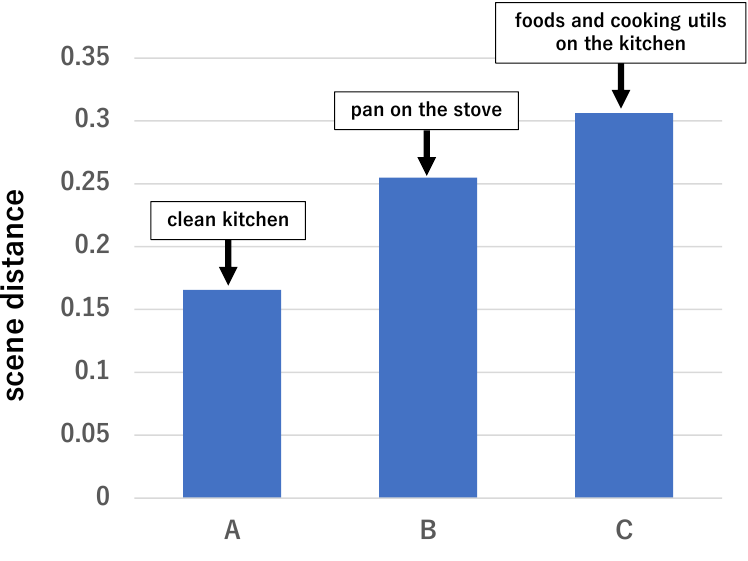}
  \caption{Scene distance and feature of three types of scenes at kitchen spot. The scene distance of image A, which is almost the same situation as the reference, is small, and the scene distances of the other images change with the degree of semantic difference in the situation.}
  \label{figure:kitchen-results}
\end{figure}

\begin{figure}[tb]
  \centering
  \includegraphics[width=0.9\columnwidth]{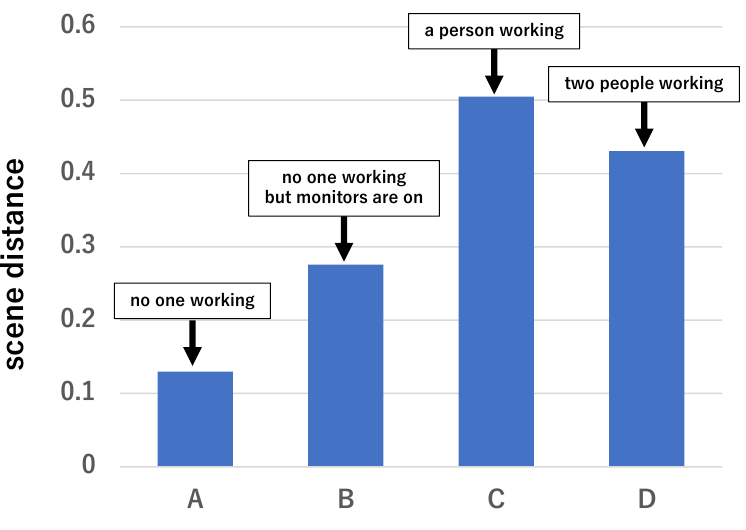}
  \caption{Scene distance and feature of four types of scenes at desk spot. The scene distance of image A, which is almost the same situation as the reference, is small, and the scene distances of the other images change with the degree of semantic difference in the situation.}
  \label{figure:desk-results}
\end{figure}

\begin{figure}[tb]
  \centering
  \includegraphics[width=0.9\columnwidth]{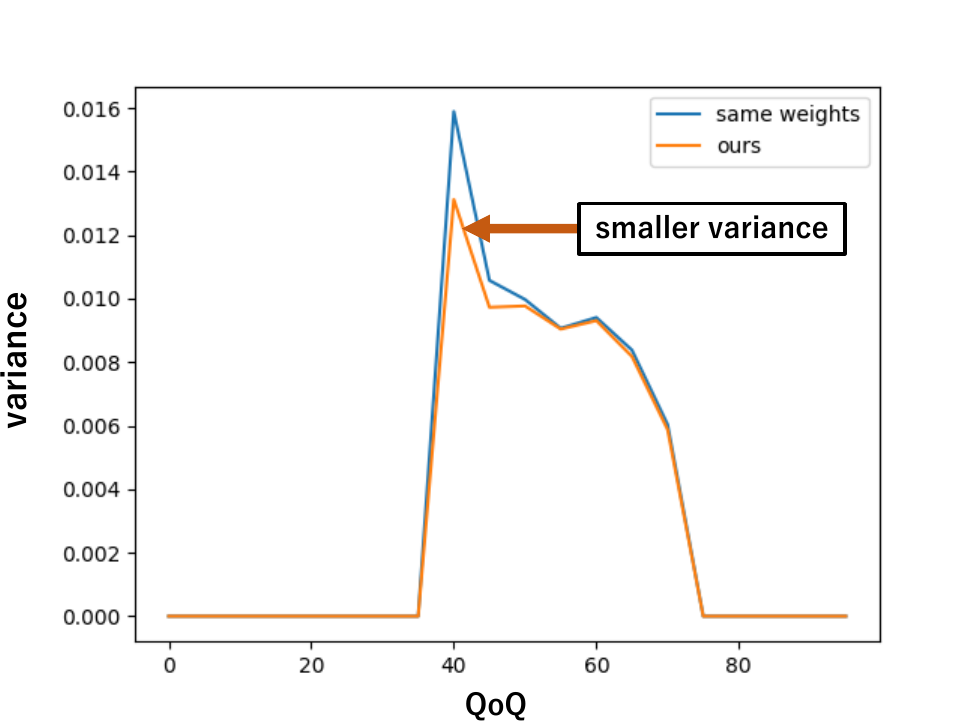}
  \caption{Comparison of variance of scene distance in C in table spot for same weights and suggested methods. Our method has a smaller variance regardless of QoQ.}
  \label{figure:table-var}
\end{figure}
\begin{figure}[tb]
  \centering
  \includegraphics[width=0.9\columnwidth]{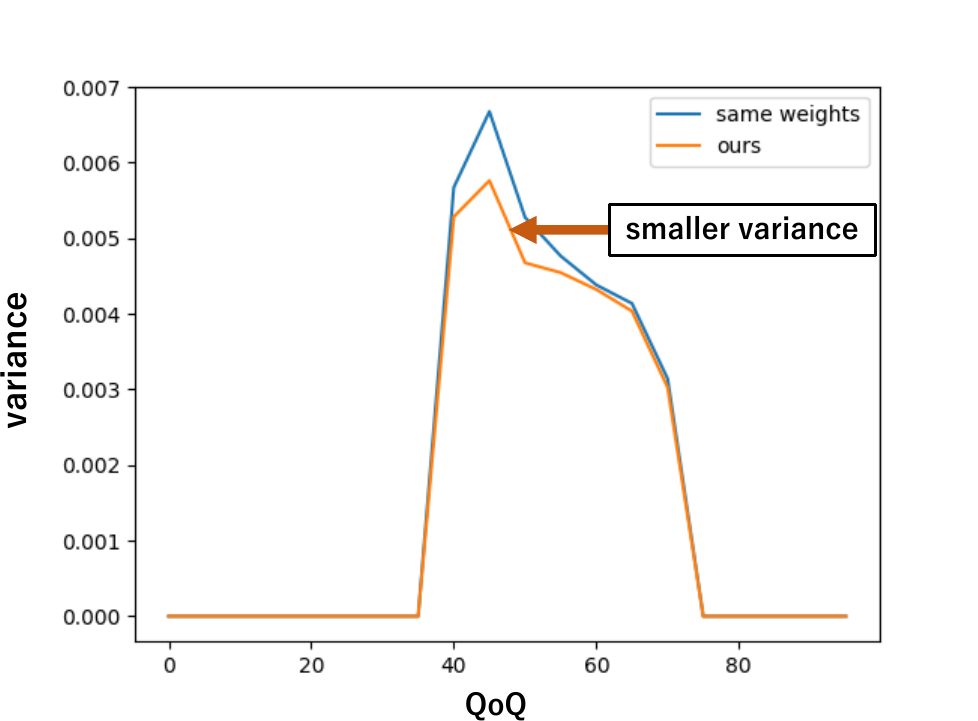}
  \caption{Comparison of variance of scene distance in C in kitchen spot for same weights and suggested methods. Our method has a smaller variance regardless of QoQ.}
  \label{figure:kitchen-var}
\end{figure}
\begin{figure}[tb]
  \centering
  \includegraphics[width=0.9\columnwidth]{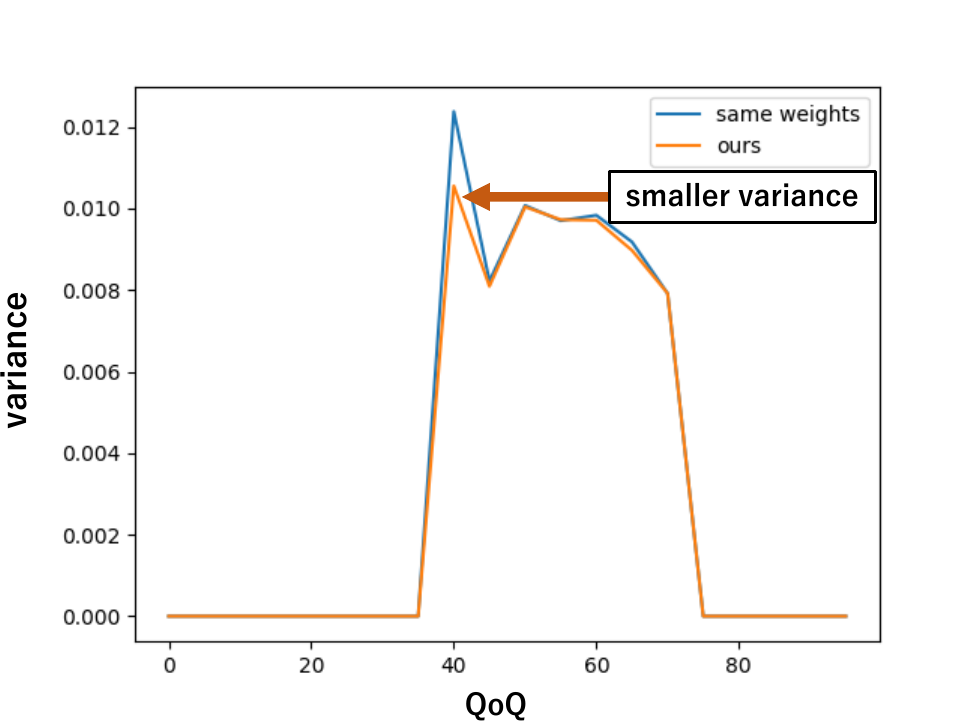}
  \caption{Comparison of variance of scene distance in C in desk spot for same weights and suggested methods. Our method has a smaller variance regardless of QoQ.}
  \label{figure:desk-var}
\end{figure}

Scene difference detection experiments were conducted in the environment of the 73B2 laboratory in Engineering Bldg. 2 at the University of Tokyo. As shown in \figref{demo-place}, this facility is a laboratory for robots, but it is also equipped with a kitchen, dining table, and work desk, making it suitable for evaluating the proposed method in multiple scenes. We use Fetch Mobile Manipulator\cite{wise2016fetch} from Fetch Robotics to move to the table spot, kitchen spot, and desk spot where demand for life support behaviors is likely to be high in the figure, move the head camera to a specific position, and capture the scene. The coordinates for capturing these images are shown in \tabref{coords}, and the angles of the torso and head are shown in \tabref{pos}. In this experiment, the robot navigated to these spots and moved the height of the torso and the angle of the head camera to the target position and angle each time, so there were slight differences in position and posture each time. We performed this imaging four times a day for two months and collected 145 images at the table spot, 144 at the kitchen spot, and 141 at the desk spot. The images used below are representative images extracted from this data.

We prepared 67 questions to express the situation. The questions include
\begin{itemize}
  \item what does this image describe?
  \item what is being done?
  \item what objects are seen?
  \item what is on it?
  \item how many people?
\end{itemize}

\subsection{Scene Distance Measurement Using All Questions}
\label{exp:all}

The spot shown in \figref{demo-place} was imaged and the scene distance was quantified with the proposed method using all 67 questions. The reference images at the table spot, kitchen spot, and desk spot and the images taken at other times are shown in \figref{experiments-photos}. The graphs of the scene distance at that time are shown in \figref{table-results}, \figref{kitchen-results} and \figref{desk-results}, respectively.

In table spot A, the placement of the remote control and keyboard is slightly different but almost the same as in Normal; in B, a cardboard box and a bag are placed; in C, a person is working on a PC; and in D, food containers are scattered around. C had the highest scene distance.

In kitchen spot, A has the kettle on the induction stove shifted to the next position; B has a pot; C has food and cooking utensils on the sink and a frying pan on the induction stove. C had the highest scene distance.

In the desk spot, A is in the same state as Normal; B has a laptop PC with a monitor; C has a person working; D has two people working. C had the highest scene distance.

\subsection{Evaluation of Variance Values for Two Different Weighting Methods}

We evaluated the weighting method proposed in Sec.\ref{subsec:evalanomaly} for calculating the scene distance at the largest scene distance for each spot. 10 questions are randomly selected from 67 questions, and 10,000 sets are created. For each set of questions, $QoQ$ of the questions is calculated. The horizontal axis represents the $QoQ$, and the vertical axis represents the variance of the scene distance for each separation bin, as plotted in \figref{table-var}，\figref{kitchen-var} and \figref{desk-var}. Both the methods Eq.(\ref{avg}) and  Eq.(\ref{ours}) in Sec.\ref{subsec:evalanomaly} are plotted. The variance of both methods decreases as $QoQ$ increases and indicates that the variance of the proposed method is smaller in regions where $QoQ$ is small.

\section{DISCUSSIONS}
\subsection{Qualitative Evaluation of Scene Distance}
We discuss the results of \figref{table-results}, \figref{kitchen-results} and \figref{desk-results}. In the table spot A and the kitchen spot A, the scene distances were less than 0.2 for the movement of the remote controller and the movement of the kettle, and the scene distances were not high for slight movement of the objects. The scene distance of A in the desk spot is about 0.14, and the change in the scene distance is considered to be small in relation to the robot's navigation and head posture deviation. The scene distance of C in the kitchen spot is larger than that of B, indicating that the scene distance can describe the amount of things and the complexity of the situation. The scene distance of C in the desk spot is higher than that of B. It is clear that the scene distance increases when the complexity of the situation where the objects in the image remain unchanged and a person starts working is added. The above results suggest that, with appropriate thresholding, it is possible to detect semantic anomaly.

\subsection{Variation in Scene Distance due to Question Bias}
We discuss the results of \figref{table-var}, \figref{kitchen-var} and \figref{desk-var}．Depending on how the questions are selected, the scene distance will have a wide range of values: for a small $QoQ$, the variance tends to be large, and for a larger $QoQ$, the variance tends to be small. This is because when $QoQ$ is small, either all the questions are close in meaning, or there are a small number of questions with different meanings mixed in with the close meaning questions, and the output similarity is affected by the close meaning questions. The proposed method achieves a smaller variance when $QoQ$ is small than when the same weights are used. This is because the contribution of a small number of questions with different meanings is increased and the influence of many questions with the same meaning is decreased. The effect of this variance suppression varies depending on the scene conditions. This is thought to be due to the influence of the similarity of the output responses, and it is necessary to examine the contribution of these responses to each other.

\subsection{Variation in Scene Distance due to Question Content}
Scene Distance varies depending on the questions used. In the experiment, we prepared many variations of the questions to capture a wide range of semantic differences. The questions used can be customized for each task. For example, if the user wants to classify the state of a person, it is possible to input questions focusing on the presence or movement of the person. In the future, when performing clustering and anomaly detection in daily life environments, the proposed system will allow users to specify clusters they wish to classify and anomalies they wish to detect using natural language questions.

\section{CONCLUSIONS}
In this study, we proposed a scene difference detection method using a pre-trained large-scale vision-language model for mobile robots to detect changes in our daily life environment. Multiple questions are prepared in advance, and reference images and current images are input to the model together with the question sentences, and the current difference is calculated by quantifying the difference between each answer sentence. In this process, the similarity of the meanings of the questions is taken into account, and the scene distance is calculated by weighting the value of the distance between the answers. Experimental results using these methods showed that, first, the semantic scene distance of the situation can be quantified and that the contribution of weak changes in the object, robot navigation, and posture errors to the scene distance is small. Second, the proposed weighting method was found to reduce the variance of the scene distance due to differences in the meaning of the selected questions. This method eliminates training costs in the local environment and may accelerate the spread of robots that can detect any semantic change in the living environment. The proposed method can be directly applied to a system in which a robot spontaneously decides whether or not to help a person by using questions about the person's actions or decides whether or not to clean up a room by using questions about the state of the objects in the room based on the normal state of its environment.

As a prospect of this research, it has the potential not only to quantify scene distances from the obtained answer sentences but also to explain changes using a spoken language with a large-scale language model such as GPT-3\cite{brown2020language}. In addition, constructing a system that clusters environmental conditions based on the obtained scene distance and unsupervised learning may also allow the robot to perform accurate tasks based on the conditions.

\addtolength{\textheight}{-12cm}   % This command serves to balance the column lengths
                                  % on the last page of the document manually. It shortens
                                  % the textheight of the last page by a suitable amount.
                                  % This command does not take effect until the next page
                                  % so it should come on the page before the last. Make
                                  % sure that you do not shorten the textheight too much.

\bibliographystyle{junsrt}
\bibliography{main}

\end{document}